\documentclass[journal]{IEEEtran}

\usepackage[numbers,sort&compress]{natbib}

\usepackage{algorithmicx}
\usepackage{algpseudocode}
\usepackage{enumerate}
\usepackage{multirow}
\usepackage{colortbl}
\usepackage{color}
\usepackage[table,xcdraw]{xcolor}
\usepackage{array}
\definecolor{mygray}{gray}{.6}
\usepackage{graphicx}
\usepackage{subfigure}
\usepackage{amsmath} 
\usepackage{float}  
\floatname{algorithm}{Algorithm} 
\usepackage{amsthm,amsmath,amssymb}
\usepackage[ruled,linesnumbered]{algorithm2e}
\usepackage{mathrsfs}
\usepackage{CJKutf8}
\usepackage{booktabs} 

\ifCLASSINFOpdf

\else

\fi

\begin{document}
\title{A Two-stage Evolutionary Framework For Multi-objective Optimization}

\author{Peng Chen,
    Jing Liang*,
	Kangjia Qiao, 
	Ponnuthurai Nagaratnam Suganthan, Xuanxuan Ban}



\maketitle	
\pagestyle{empty}  
\thispagestyle{empty} 

\begin{abstract}
  In the field of evolutionary multi-objective optimization, the approximation of the Pareto front (PF) is achieved by utilizing a collection of representative candidate solutions that exhibit desirable convergence and diversity. Although several multi-objective evolutionary algorithms (MOEAs) have been designed, they still have difficulties in keeping balance between convergence and diversity of population. To better solve multi-objective optimization problems (MOPs), this paper proposes a Two-stage Evolutionary Framework For Multi-objective Optimization (TEMOF). Literally, algorithms are divided into two stages to enhance the search capability of the population. During the initial half of evolutions, parental selection is exclusively conducted from the primary population. Additionally, we not only perform environmental selection on the current population, but we also establish an external archive to store individuals situated on the first PF. Subsequently, in the second stage, parents are randomly chosen either from the population or the archive. In the experiments, one classic MOEA and two state-of-the-art MOEAs are integrated into the framework to form three new algorithms. The experimental results demonstrate the superior and robust performance of the proposed framework across a wide range of MOPs. Besides, the winner among three new algorithms is compared with several existing MOEAs and shows better results. Meanwhile, we conclude the reasons that why the two-stage framework is effect for the existing benchmark functions.
\end{abstract}


\IEEEpeerreviewmaketitle

\section{Introduction}

Multi-objective optimization problems (MOPs) \cite{deb2016multi,pereira2022review,sharma2022comprehensive} frequently entail the optimization of two or more conflicting objectives simultaneously, a scenario commonly encountered in real-world applications. As it is improbable to identify a single solution that optimizes all conflicting objectives, it is typically expected to obtain a range of solutions that represent trade-offs between different objectives. These trade-off solutions, known as Pareto optimal solutions, collectively form the Pareto set (PS), and the visualization of the PS in objective space is referred to as the Pareto front (PF).

Over the past two decades, the multi-objective evolutionary algorithms (MOEAs) have exhibited significant efficacy in addressing MOPs \cite{zhou2011multiobjective}. In relation to the environmental selection strategies, the current MOEAs can be broadly categorized into three groups, namely Pareto dominance based MOEAs, decomposition based MOEAs, and indicator based MOEAs. The Pareto dominance based MOEAs utilize non-dominated sorting approaches \cite{zhang2014efficient} to partition candidate solutions into multiple ranks initially, and subsequently differentiate the candidate solutions within the same rank using other diversity metrics \cite{deb2002fast,he2017radial}. The decomposition based MOEAs are characterized by breaking down the original MOP into a series of single-objective optimization problems (SOPs) or simpler MOPs to be collaboratively solved \cite{zhang2007moea,deb2013evolutionary}. Regarding the indicator based MOEAs, where the environmental selection is dependent on a performance indicator such as hypervolume (HV) \cite{while2006faster}, inverted generational distance \cite{zhou2006combining}, and R2 \cite{trautmann2013r2}, the fitness of a candidate solution is assessed based on its contribution to the indicator value in relation to the entire population \cite{bader2011hype,hernandez2015improved}.

Although numerous MOEAs have been developed, they lack the capability to effectively filter and preserve high-quality solutions. These algorithms struggle to strike a balance between the diversity and convergence of the population across different search periods. Consequently, they typically require more computational resources to explore the global optimum and are prone to getting trapped in local optima. Hence, we devise a novel Two-stage Evolutionary Framework For Multi-objective Optimization (TEMOF). This framework consists of two distinct stages. During the initial half of evolutions, parents are exclusively selected from the primary population. In the subsequent stage, parents are randomly chosen either from the main population or the archive, based on the parameter $p$, thereby enhancing the algorithm's convergence capability. In the conducted experiments, three novel algorithms are introduced by employing the classic NSGA-III \cite{deb2013evolutionary} and the two advanced MOEAs \cite{tian2018guiding,li2015pareto}. These algorithms are collectively called primitive MOEAs (PMOEAs) in this paper. The outcomes from the experiments effectively validate the efficacy of the framework on a diverse range of benchmark problems.

The rest of this paper is arranged as follows. Section \ref{Background} introduces the Background. Section \ref{The Proposed Framework} and \ref{Ex} describe the proposed framework and experimental results respectively. Section \ref{Conclusion} concludes this paper and gives the future work.

\section{Background}
\label{Background}
\subsection{Related Work}

In this section, we provide a concise overview of selection strategies in MOEAs, categorizing them into four primary groups \cite{zhou2023evolutionary}. 

1) The first group is centered around the decomposition structure, exemplified by selection strategies employed in MOEA/D \cite{zhang2007moea}, NSGA-III \cite{deb2013evolutionary}, and MOEA/DD \cite{li2014evolutionary}. In MOEA/D, a selection strategy based on weight vector and aggregation scheme was introduced to uphold population diversity and convergence. NSGA-III integrated a selection strategy involving Pareto optimality, coupled with a decomposition-based diversity preservation operation. MOEA/DD devised a selection strategy that combines Pareto optimality and decomposition structure, aiming to strike a balance between convergence and diversity during evolution. However, a notable drawback of these strategies lies in the challenge of determining appropriate weight vectors. Furthermore, the determination of the neighborhood size in these strategies is crucial for achieving a balance between convergence and diversity, presenting a non-trivial task. Additionally, the effectiveness of these strategies is contingent upon the choice of the aggregation method \cite{das1998normal}.

2) The second category revolves around a modified Pareto dominance approach, encompassing selection strategies found in GrEA \cite{yang2013grid} and $\epsilon$-MOEA \cite{mohan2005evaluating}. In $\epsilon$-MOEA, the selection scheme is constructed based on the concept of $\epsilon$-dominance. This scheme employs archive updating strategies to select a distributed and converged set of solutions. Meanwhile, GrEA's selection is formulated on grid dominance and difference, delineating the relationship among individuals within the grid. Both selection strategies involve parameters (such as hyper-box size in $\epsilon$-MOEA and grid size in GrEA) that pose challenges in proper configuration.

3) The third category of selection strategies tends to leverage fitness assignment methods, incorporating approaches such as quality indicators (e.g., hypervolume indicator in HypE \cite{bader2011hype}), predefined target points (e.g., knee point in KnEA \cite{zhang2014knee}), or preference indicators (e.g., preference indicators in PICEA-g \cite{wang2012preference}). These methods are utilized to assign fitness values to solutions, reflecting their convergence and diversity performance. Subsequently, selection is executed based on the distances between each solution and the predefined target points. As highlighted in \cite{yen2013performance}, it is acknowledged that no indicator can perfectly assign fitness values to each solution; rather, they can only offer specific yet incomplete quantifications of solutions. Additionally, since the selection criterion is grounded in the fitness of solutions, there is a heightened likelihood that individuals with higher fitness will be consistently chosen for combination, potentially resulting in a loss of diversity.

 4) The last group of selection strategy is associated with co-evolution based MOEAs. For instance, archives are utilized in the co-evolutionary-constrained multi-objective optimization algorithm (CCMO) \cite{tian2020coevolutionary} and the dual-population-based evolutionary algorithm for constrained multiobjective optimization (c-DPEA) \cite{ming2021dual} in order to strike a balance between constraints and the quality of exploration. Furthermore, archives are also utilized in CPDEA \cite{liu2019handling} and MMEA-WI \cite{li2021weighted} to effectively manage the trade-off between diversity and convergence. In addition, a novel co-evolutionary framework called multiple populations for multi-objective optimization (MPMO) is designed to enhance the performance of MOEAs. Then, numerous variants of MPMO emerged. A co-evolutionary multi-swarm PSO (CMPSO) \cite{zhan2013multiple} is designed based on the MPMO framework, with the extra use of an external shared archive for information exchange between different sub-populations. After that, inspired by CMPSO, CMODE \cite{wang2015cooperative} is proposed based on the MPMO framework by using a co-operative differential evolution (DE) \cite{cui2016adaptive} with multiple sub-populations to optimize multiple objectives. Each sub-population optimizes only one objective and all the sub-populations cooperate to approximate the whole PF using the DE operator.

\subsection{Indicator}

There are a large number of performance metrics for evaluating the quality of solution set obtained by MOEAs, which can be roughly divided into the following two categories \cite{tian2016multi}.

For the first category, priori knowledge about the Pareto optimal front is required in the metric calculations. Two representative metrics of this category are the IGD \cite{zhou2006combining} and the generational distance (GD) \cite{van1998multiobjective}, both of which require a set of uniformly distributed reference points sampled from the Pareto optimal front as priori knowledge. The incorporation of both convergence accuracy and speed within the U-score approach \cite{price2023trial} provides a comprehensive perspective on algorithmic performance, thereby facilitating effective comparative analyses and rankings across a multitude of algorithms by considering each run of each algorithm.

For the second category, there is no priori knowledge about the Pareto optimal front required in the metric calculations. The most representative metric of this type is the HV indicator \cite{while2006faster}. The calculation of HV only requires a set of predefined reference points. A higher HV value indicates the better convergence as well as diversity of the points in Population.

\section{The Proposed Framework}
\label{The Proposed Framework}
The pseudo-code of TEMOF is elucidated in Algorithm \ref{algorithm1}. In accordance with the majority of MOEAs, TEMOF is comprised of the subsequent components: population initialization, mating selection, offspring generation, and environmental selection. 


     \begin{algorithm*}[h]
 	\caption{Proposed TEMOF Framework} \label{algorithm1}
 	\SetKwInOut{Input}{Input}\SetKwInOut{Output}{Output}
 	
 	\Input{$N$ (population size) \\ $maxFEs$ (maximum number of evaluation resources) \\ }
 	\Output{$Archive$ (Convergence archive) \\}
 	 $Population \Leftarrow Initialization(N) $\;
 	 Parameter initialization // Updates to parameters for PMOEAs\;
 	 $Archive$ = $Population$ \;
 	 $FEs$ = $N$ \;
 	 \While{$FEs \leq  maxFEs$} {

 	\eIf{$FEs \geq 0.5*maxFEs and rand < p$}{
 	{$Off \Leftarrow \pmb{ OffspringGeneration} (Archive, N)$ // Generating offspring using the Archive;}
 	}{
 	{$Off \Leftarrow \pmb{ OffspringGeneration} (Population, N)$ // Generating offspring using the Population;}
 	}
         $Pop \Leftarrow \pmb{ EnvironmentalSelection} ([Population, Off], N)$ // Simultaneously preserving individuals with good diversity or convergence characteristics\;
         $Archive \Leftarrow \pmb{ EnvironmentalSelection2} ([Archive, Off], N)$ // Based on EnvironmentalSelection, only individuals located on the first level of the PF are preserved\;
         $Arc \Leftarrow [Pop, Archive]$   // Merge two sets and remove duplicate elements \;
         $Population \Leftarrow \pmb{ EnvironmentalSelection} (Arc, N)$ \;

}
 	 return $Population$ \;
    
 \end{algorithm*}

 Firstly, the population, parameters, and archive are initialized. The parameters represent inherent settings for PMOEAs. Subsequently, we conduct a phased search until reaching the termination condition. It is worth noting that a convergence archive to ameliorate the quality of convergence and to regulate the caliber of the acquired local PFs is introduced. In each generation, the framework opts for parents from both the main population and the convergence archive to execute co-evolution. This paper suggests setting the parameter $p$ to $0.5$. The methods for offspring generation and environmental selection directly follow PMOEAs. Specifically, EnvironmentalSelection2 is an extension of EnvironmentalSelection, preserving individuals only from the front ranked as the first level.

Please note that the proposed framework is simple and easy to be implemented. In the following section, three representative PMOEAs will be embedded into this framework to form three new algorithms. Due to page limit, the pseudo-codes of three new algorithms do not introduce in this paper.

\section{Experiments}
\label{Ex}
\subsection{Experimental Settings}

1) Benchmark functions: In accordance with MOPs within CEC 2024 competition and special session on numerical optimization, we employ the MaOPs \cite{li2019comparison}, comprising 10 multi-objective problems, to assess and rank the optimizers. The maximum number of evaluations is set at 100,000 for each function. According to the recommendation, both the population size and the final Pareto Front (Front 1) size are set to 100. Furthermore, the decision space dimension ($M$) and the objective space dimension ($D$) for each test problem are uniformly set to $3$ and $7$, respectively.

2) Base algorithms and compared algorithms: The classic NSGA-III \cite{deb2013evolutionary} and the two advanced MOEAs \cite{tian2018guiding,li2015pareto} are adopted to form three new algorithms to verify the effectiveness of the proposed framework. These three algorithms are designated as TEMOF-NSGA-III, TEMOF-GFMMOEA, and TEMOF-DWU, respectively. In addition, the superior algorithm among the three new ones will be compared with other MOEAs including \cite{tian2017indicator,moreira2019guiding,jiang2017strength}. Parameters of all algorithms keep unchanged. Please note that all experiments are implemented on PlatEMO \cite{tian2017platemo}.

3) Performance metrics: In this paper, two quality metrics, IGD and HV, are primarily utilized. In addition, the Wilcoxon’s rank-sum test at a 0.05 significance level is used to compare two algorithms based on the performance indicators. Symbols ``+'', ``-'', and ``='' show that the compared algorithm is better than, worse than, and equal to the base algorithm.

\begin{table*}[!htpb]
  \centering
  \caption{Comparison Results of IGD among TEMOF-MOEAs}
  \setlength{\tabcolsep}{14pt}
\begin{tabular}{cccccc}
\hline
Problem      & TEMOF-BCEIBEA                                & TEMOF-GFMMOEA         & TEMOF-NSGA-III                             \\
\hline
MaOP1        & \pmb{ 1.4e+1 (1.0e+0) =} & 1.5e+1 (2.0e+0) - & 1.4e+1 (7.6e-1)                        \\
MaOP2        & 2.9e+0 (2.4e+0) =                        & 3.3e+1 (2.4e+1) - & \pmb{ 2.4e+0 (1.6e+0)} \\
MaOP3        & \pmb{ 1.6e+1 (1.2e+0) =} & 1.7e+1 (1.2e+0) = & 1.7e+1 (6.1e-1)                        \\
MaOP4        & 4.6e-1 (1.2e-1) -                        & 5.3e-1 (1.9e-1) - & \pmb{ 4.2e-1 (5.9e-2)} \\
MaOP5        & 1.1e+0 (4.7e-1) -                        & 2.2e+0 (1.2e+0) - & \pmb{ 9.5e-1 (3.5e-1)} \\
MaOP6        & \pmb{ 8.7e-1 (2.5e-1) +} & 3.1e+0 (3.8e+0) - & 1.0e+0 (2.5e-1)                        \\
MaOP7        & 5.2e+0 (3.9e+0) =                        & 6.1e+0 (4.7e+0) = & \pmb{ 4.5e+0 (3.5e+0)} \\
MaOP8        & 6.0e+0 (4.7e+0) -                        & 6.2e+0 (5.2e+0) - & \pmb{ 3.5e+0 (3.0e+0)} \\
MaOP9        & 6.4e+0 (4.9e+0) -                        & 6.4e+0 (6.1e+0) - & \pmb{ 4.6e+0 (2.9e+0)} \\
MaOP10       & 6.7e+0 (7.4e+0) -                        & 7.2e+0 (8.1e+0) - & \pmb{ 5.1e+0 (2.8e+0)} \\
\midrule
+/-/= & 1/5/4                 & 0/8/2                 &           \\                              \hline
\end{tabular}
\label{TEMOF-result}
\end{table*}

\begin{table*}[!htpb]
  \centering
  \caption{Comparison Results of HV among TEMOF-MOEAs}
  \setlength{\tabcolsep}{14pt}
\begin{tabular}{cccccc}
\hline
Problem      & TEMOF-BCEIBEA                                & TEMOF-GFMMOEA         & TEMOF-NSGA-III                             \\
\hline
MaOP1   & 8.3e-1 (3.4e-2) - & 6.4e-1 (9.1e-2) - & \pmb{ 9.1e-1 (2.5e-2)} \\
MaOP2   & 1.3e+0 (2.3e-5) = & 1.3e+0 (1.1e-2) - & \pmb{ 1.3e+0 (9.5e-6)} \\
MaOP3   & 1.0e+0 (4.0e-2) - & 1.0e+0 (3.0e-2) - & \pmb{ 1.0e+0 (1.8e-2)} \\
MaOP4   & 1.3e+0 (4.3e-3) = & 1.3e+0 (1.7e-2) = & \pmb{ 1.3e+0 (2.5e-3)} \\
MaOP5   & 1.3e+0 (1.0e-3) = & 1.3e+0 (2.0e-2) - & \pmb{ 1.3e+0 (6.3e-4)} \\
MaOP6   & 1.3e+0 (4.6e-3) = & 1.3e+0 (7.3e-3) - & \pmb{ 1.3e+0 (3.5e-4)} \\
MaOP7   & 1.2e+0 (8.6e-2) - & 1.0e+0 (1.4e-1) - & \pmb{ 1.3e+0 (1.1e-2)} \\
MaOP8   & 1.2e+0 (6.7e-2) - & 1.2e+0 (7.9e-2) - & \pmb{ 1.3e+0 (1.1e-2)} \\
MaOP9   & 1.2e+0 (5.4e-2) - & 1.1e+0 (7.4e-2) - & \pmb{ 1.3e+0 (5.7e-3)} \\
MaOP10  & 1.2e+0 (1.2e-1) - & 1.2e+0 (6.2e-2) - & \pmb{ 1.3e+0 (1.6e-2)} \\
\midrule
+/-/=   & 0/6/4                 & 0/9/1                 &                                                      \\              
\hline
\end{tabular}
\label{TEMOF-result-HV}
\end{table*}

\begin{table*}[!]
  \centering
  \caption{Results of IGD obtained by the Wilcoxon test for algorithm TEMOF-NSGA-III against TEMOF-MOEAs}
  \setlength{\tabcolsep}{14pt}
\centering\small
\begin{tabular}{ccccc}
\hline
\multicolumn{5}{c}{{IGD}}\\
\hline
 TEMOF-NSGA-III VS. & $R^{+}$ & $R^{-}$  &  $P-$value & $\alpha$ $\le$ 0.05\\ \hline 
TEMOF-BCEIBEA & 44.0 & 11.0 & 0.083131  & No\\ \hline 
TEMOF-GFMMOEA & 55.0 & 0.0  & 0.004317  & Yes\\ \hline 
\multicolumn{5}{c}{{HV}}\\
\hline
 TEMOF-NSGA-III VS. & $R^{+}$ & $R^{-}$  &  $P-$value & $\alpha$ $\le$ 0.05\\ \hline 
TEMOF-BCEIBEA & 50.0 & 5.0  & 0.017566 & Yes\\ \hline 
TEMOF-GFMMOEA & 55.0 & 0.0 & 0.000903 & Yes \\ \hline 

    \end{tabular}%
  \label{Results obtained by the Wilcoxon test for algorithm TEMOF-NSGA against TEMOF-MOEAs}%
\end{table*}%

\begin{figure}[!h]
    \centering
    \includegraphics[width=10cm]{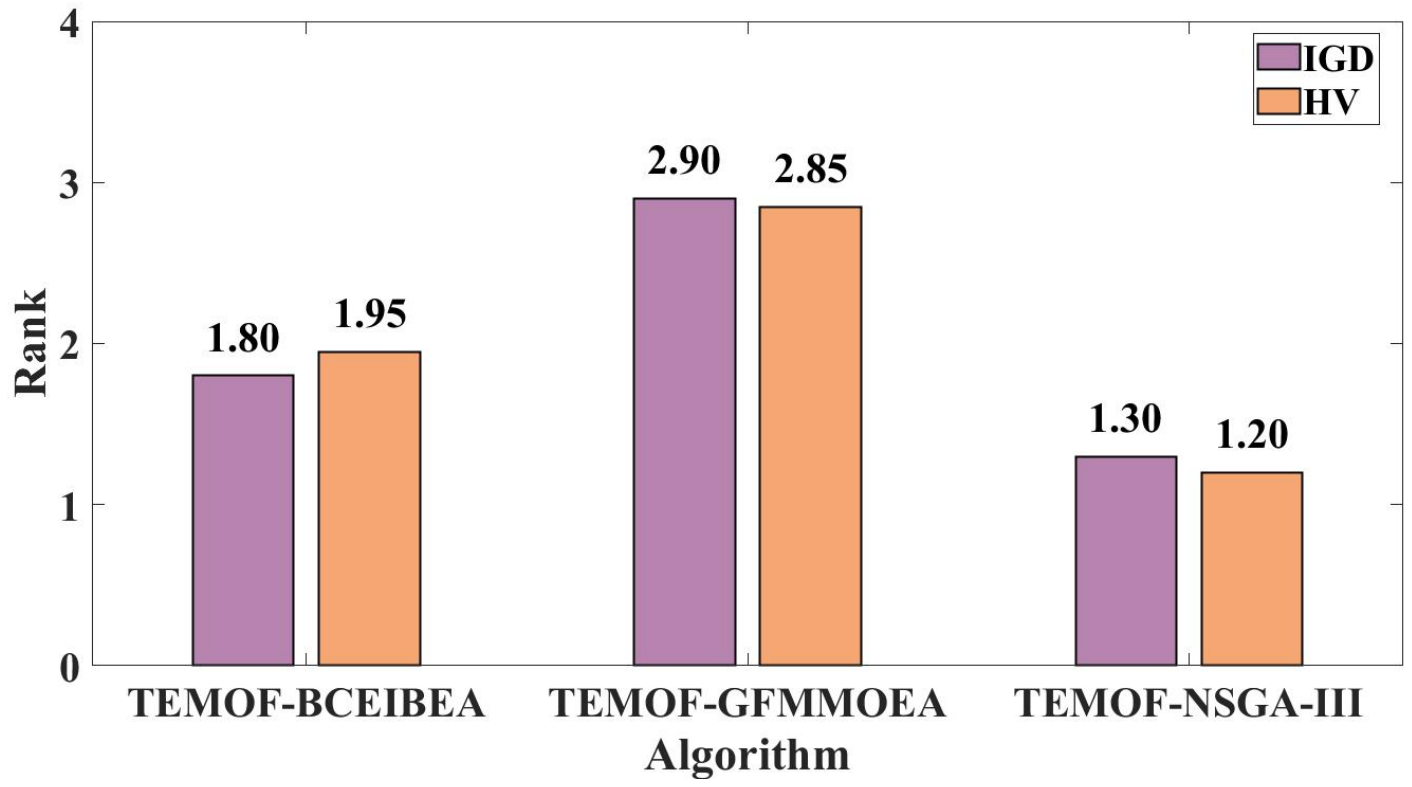}
    \caption{Fridman Ranking of TEMOF-MOEAs}
    \label{Fridman1}
\end{figure}

\begin{table*}[!ht]
  \centering
  \caption{Comparison Results of IGD among TEMOF-NSGA-III and MOEAs}
  \setlength{\tabcolsep}{0.5pt}
\begin{tabular}{cccccccccc}
\hline
Problem       & ARMOEA                                       & BCEIBEA                                      & DWU                   & GFMMOEA               & NSGA-III                                      & SPEAR                 & TEMOF-NSGA-III                             \\
\hline
MaOP1        & 1.4e+1 (1.7e+0) =                        & \pmb{ 1.3e+1 (1.3e+0) +} & 2.0e+1 (1.9e+0) - & 1.5e+1 (1.3e+0) = & 1.5e+1 (1.6e+0) =                        & 1.9e+1 (3.4e+0) - & 1.4e+1 (7.6e-1)                        \\
MaOP2        & 3.6e+0 (4.0e+0) =                        & \pmb{ 2.3e+0 (2.0e+0) =} & 1.3e+2 (2.7e+1) - & 3.0e+1 (1.7e+1) - & 2.7e+0 (2.3e+0) =                        & 3.7e+0 (5.5e+0) = & 2.4e+0 (1.6e+0)                        \\
MaOP3        & 1.7e+1 (1.0e+0) -                        & 1.7e+1 (7.1e-1) =                        & 1.9e+1 (9.6e-1) - & 1.8e+1 (1.1e+0) - & \pmb{ 1.6e+1 (1.2e+0) +} & 1.8e+1 (1.0e+0) - & 1.7e+1 (6.1e-1)                        \\
MaOP4        & 8.8e-1 (9.7e-1) -                        & 5.2e-1 (2.5e-1) =                        & 4.8e-1 (2.7e-2) - & 4.5e-1 (1.3e-1) - & 4.5e-1 (1.3e-1) =                        & 7.4e-1 (2.1e-1) - & \pmb{ 4.2e-1 (5.9e-2)} \\
MaOP5        & 1.2e+0 (5.8e-1) -                        & 1.1e+0 (5.9e-1) =                        & 3.9e+0 (3.4e+0) - & 2.2e+0 (1.6e+0) - & \pmb{ 8.1e-1 (5.7e-1) +} & 2.0e+0 (6.3e-1) - & 9.5e-1 (3.5e-1)                        \\
MaOP6        & \pmb{ 9.2e-1 (2.6e-1) =} & 1.0e+0 (2.7e-1) =                        & 2.4e+0 (1.6e+0) - & 4.4e+0 (4.2e+0) - & 1.1e+0 (3.3e-1) =                        & 3.5e+0 (2.5e+0) - & 1.0e+0 (2.5e-1)                        \\
MaOP7        & 7.9e+0 (8.7e+0) -                        & 6.5e+0 (5.4e+0) -                        & 3.7e+1 (4.1e+1) - & 5.8e+0 (7.7e+0) = & 6.7e+0 (5.5e+0) -                        & 6.2e+1 (3.3e+1) - & \pmb{ 4.5e+0 (3.5e+0)} \\
MaOP8        & 8.3e+0 (6.1e+0) -                        & 7.9e+0 (6.0e+0) -                        & 7.2e+1 (6.7e+1) - & 7.2e+0 (5.2e+0) - & 6.7e+0 (5.5e+0) -                        & 5.0e+1 (3.5e+1) - & \pmb{ 3.5e+0 (3.0e+0)} \\
MaOP9        & 7.5e+0 (8.7e+0) -                        & 6.7e+0 (5.8e+0) -                        & 6.8e+1 (6.6e+1) - & 5.0e+0 (3.6e+0) = & 8.0e+0 (6.1e+0) -                        & 4.7e+1 (3.0e+1) - & \pmb{ 4.6e+0 (2.9e+0)} \\
MaOP10       & 7.3e+0 (9.1e+0) -                        & 8.8e+0 (5.7e+0) -                        & 9.5e+1 (6.7e+1) - & 7.2e+0 (9.6e+0) = & 8.3e+0 (5.0e+0) -                        & 4.6e+1 (3.4e+1) - & \pmb{ 5.1e+0 (2.8e+0)} \\
\midrule
+/-/= & 0/7/3                                        & 1/4/5                                        & 0/10/0                & 0/6/4                 & 2/4/4                                        & 0/9/1                 &                                      \\
\hline
\label{2}
\end{tabular}
\end{table*}

\begin{table*}[!ht]
  \centering
  \caption{Comparison Results of HV among TEMOF-NSGA-III and MOEAs}
  \setlength{\tabcolsep}{0.5pt}
\begin{tabular}{cccccccccc}
\hline
Problem       & ARMOEA                                       & BCEIBEA                                      & DWU                   & GFMMOEA               & NSGA-III                                      & SPEAR                 & TEMOF-NSGA-III                             \\
\hline
MaOP1                       & 8.0e-1 (9.4e-2) - & 8.1e-1 (4.1e-2) - & 7.9e-1 (3.8e-2) - & 7.0e-1 (9.5e-2) -                        & 8.7e-1 (4.3e-2) -                        & 7.1e-1 (8.5e-2) - & \pmb{ 9.1e-1 (2.5e-2)} \\
MaOP2                       & 1.3e+0 (7.4e-4) - & 1.3e+0 (6.3e-5) = & 1.2e+0 (2.6e-2) - & 1.3e+0 (1.1e-2) -                        & 1.3e+0 (3.3e-5) =                        & 1.3e+ (2.4e-3) - & \pmb{ 1.3e+0 (9.5e-6)} \\
MaOP3                       & 9.9e-1 (3.5e-2) - & 1.0e+0 (2.5e-2) - & 8.7e-1 (1.0e-1) - & 9.8e-1 (6.7e-2) -                        & \pmb{ 1.0e+0 (1.9e-2) =} & 7.6e-1 (1.2e-1) - & 1.0e+0 (1.8e-2)                        \\
MaOP4                       & 1.3e+0 (7.6e-3) - & 1.3e+0 (7.0e-4) = & 1.3e+0 (4.5e-7) = & \pmb{ 1.3e+0 (1.7e-7) =} & 1.3e+0 (6.7e-3) -                        & 1.3e+0 (9.3e-3) - & 1.3e+0 (2.5e-3)                        \\
MaOP5                       & 1.3e+0 (7.2e-3) - & 1.3e+0 (2.7e-3) - & 1.3e+0 (1.6e-2) - & 1.3e+0 (4.5e-3) -                        & \pmb{ 1.3e+0 (8.2e-5) =} & 1.3e+0 (5.3e-3) - & 1.3e+0 (6.3e-4)                        \\
MaOP6                       & 1.3e+0 (4.8e-3) - & 1.3e+0 (3.1e-3) = & 1.3e+0 (1.3e-3) = & 1.3e+0 (3.8e-3) -                        & 1.3e+0 (6.6e-3) -                        & 1.3e+0 (3.0e-2) - & \pmb{ 1.3e+0 (3.5e-4)} \\
MaOP7                       & 1.3e+0 (1.6e-2) = & 1.3e+0 (1.5e-2) = & 1.2e+0 (9.0e-2) - & 1.3e+0 (3.5e-2) -                        & 1.2e+0 (2.1e-2) -                        & 1.1e+0 (1.1e-1) - & \pmb{ 1.3e+0 (1.1e-2)} \\
MaOP8                       & 1.3e+0 (1.3e-2) = & 1.3e+0 (1.6e-2) = & 1.1e+0 (1.5e-1) - & 1.3e+0 (2.2e-2) =                        & 1.2e+0 (2.6e-2) -                        & 1.1e+0 (1.1e-1) - & \pmb{ 1.3e+0 (1.1e-2)} \\
MaOP9                       & 1.3e+0 (2.6e-2) = & 1.3e+0 (1.2e-2) = & 1.1e+0 (1.2e-1) - & 1.3e+0 (6.8e-3) =                        & 1.3e+0 (1.5e-2) -                        & 1.2e+0 (6.0e-2) - & \pmb{ 1.3e+0 (5.7e-3)} \\
MaOP10                      & 1.2e+0 (4.1e-2) = & 1.2e+0 (2.0e-2) - & 1.0e+0 (2.1e-1) - & 1.3e+0 (2.1e-2) =                        & 1.3e+0 (1.5e-2) =                        & 1.1e+0 (1.3e-1) - & \pmb{ 1.3e+0 (1.6e-2)} \\
\midrule
\multicolumn{1}{l}{+/-/=}   & 0/6/4                 & 0/4/6                 & 0/8/2                 & 0/6/4                                        & 0/6/4                                        & 0/10/0                &                                           
     \\
\hline
\label{HVMOEA}
\end{tabular}
\end{table*}

\begin{table*}[!ht]
  \centering
  \caption{Results obtained by the Wilcoxon test for algorithm TEMOF-NSGA-III against MOEAs}
  \setlength{\tabcolsep}{14pt}
\centering\small
\begin{tabular}{ccccc}
\hline
\multicolumn{5}{c}{{IGD}}\\
\hline
 TEMOF-NSGA-III VS. & $R^{+}$ & $R^{-}$  &  $P-$value & $\alpha$ $\le$ 0.05\\ \hline 
ARMOEA & 54.0 & 1.0  & 0.005922  & Yes\\ \hline 
BCEIBEA & 46.0 & 9.0  & 0.052787 & No\\ \hline 
DWU & 55.0 & 0.0  & 0.004317 & Yes\\ \hline 
GFMMOEA & 55.0 & 0.0  & 0.004317 & Yes\\ \hline 
NSGAIII & 46.0 & 9.0  & 0.052787 & No\\ \hline 
SPEAR & 55.0 & 0.0  & 0.004317 & Yes\\ \hline 
    \hline
    \multicolumn{5}{c}{{HV}}\\
    \hline
 TEMOF-NSGA-III VS. & $R^{+}$ & $R^{-}$  &  $P-$value & $\alpha$ $\le$ 0.05\\ \hline 
ARMOEA & 43.5 & 1.5 & 0.002791 & Yes\\ \hline 
BCEIBEA & 43.5 & 1.5  & 0.005945 & Yes\\ \hline 
DWU & 53.5 & 1.5  & 0.006910 & Yes\\ \hline 
GFMMOEA & 43.5 & 1.5  & 0.002791 & Yes\\ \hline 
NSGAIII & 50.0 & 5.0  & 0.016142 & Yes\\ \hline 
SPEAR & 45.0 & 0.0  & 0.006434 & Yes\\ \hline 
 \hline
    \end{tabular}%
  \label{Results obtained by the Wilcoxon test for algorithm TEMOF-NSGA against MOEAs}%
\end{table*}%

\begin{figure*}[!h]
    \centering
    \includegraphics[width=14cm]{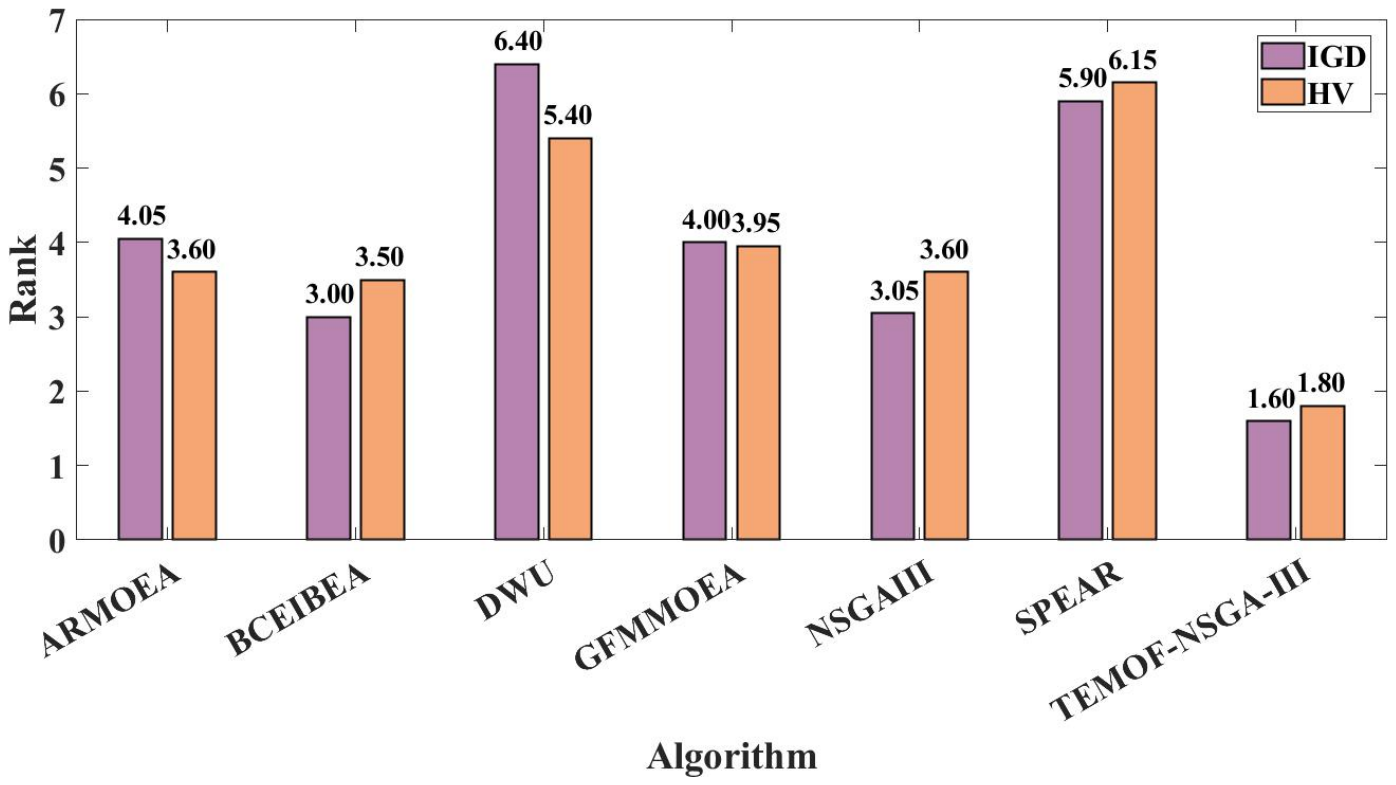}
    \caption{Fridman Ranking of Comparision Algorithms}
    \label{Fridman}
\end{figure*}

\subsection{Compared Results}

Table \ref{TEMOF-result} and Table \ref{TEMOF-result-HV} record the compared results in terms of IGD and HV respectively among improved MOEAs (TEMOF-MOEAs) on the test suit, in which the best results are marked in bold. 

 From the Tables, it can be seen that TEMOF-NSGA-III shows better performance than the other TEMOF-MOEAs. Specifically, TEMOF-NSGA-III is super than and worse than TEMOF-BCEIBEA on 5 and 1 functions in terms 
 of IGD and on 6 and 0 functions in terms 
 of HV, respectively. And TEMOF-NSGA-III is super than and worse than TEMOF-GFMMOEA on 8 and 0 functions and on 9 and 0 functions in terms of HV, separately. In addition, we compute the rankings of TEMOF-BCEIBEA, TEMOF-GFMMOEA, and TEMOF-NSGA-III, which is shown in Figure \ref{Fridman1}.

 Additionally, in terms of IGD, Table \ref{Results obtained by the Wilcoxon test for algorithm TEMOF-NSGA against TEMOF-MOEAs} indicates a significant divergence in performance between TEMOF-NSGA-III and TEMOF-GFMMOEA. While the disparity with TEMOF-BCEIBEA is comparatively minor, it still surpasses TEMOF-BCEIBEA. However, in terms of the HV, TEMOF-NSGA-III demonstrates significant differences and superiority to both TEMOF-BCEIBEA and TEMOF-GFMMOEA. Considering the comprehensive analysis, TEMOF-NSGA-III emerges as the prevailing choice among current TEMOF-MOEAs.

\subsection{Verification of the Proposed Framework}

As the victor among TEMOF-MOEAs, TEMOF-NSGA-III is compared with MOEAs, and the results are presented in Table \ref{2} and Table \ref{HVMOEA}. The compared algorithms encompass three PMOEAs and three state-of-the-art algorithms, thereby validating the efficacy of TEMOF-NSGA-III. 

From the results, it is evident that TEMOF-NSGA-III outperforms ARMOEA, DWU, GFMMOEA, and SPEAR algorithms. Significant advantages are observed in the outcomes of 7, 10, 6, and 9 out of ten problems of IGD and 6, 8, 6, 10 out of ten problems of HV, respectively. In comparison with BCEIBEA and NSGA-III, TEMOF-NSGA-III exhibits a marginal lead. It attains advantages in the results of 4 or 6 out of ten problems.

Moreover, as seen in Table \ref{Results obtained by the Wilcoxon test for algorithm TEMOF-NSGA against MOEAs}, TEMOF-NSGA-III demonstrates significant divergence and superiority in comparison with ARMOEA, DWU, GFMMOEA, and SPEAR algorithms. While the differences with BCEIBEA and NSGA-III are relatively minor, the considerably higher $R^{+}$ compared to $R^{-}$ suggests TEMOF-NSGA-III's clear superiority over these two algorithms. Additionally, from Figure \ref{Fridman}, it is evident that the ranking of results for TEMOF-NSGA-III is significantly ahead of other algorithms.

\subsection{Discussion}

Through all the conducted experiments, it can be concluded that the proposed TEMOF-NSGA-III outperforms other algorithms on multiple test problems. In addition to the original features of NSGA-III, the algorithm's staged approach contributes to balancing convergence and diversity. Furthermore, maintaining an external archive of non-dominated solutions allows the algorithm to preserve and advance multiple effective solutions. This enhances the algorithm's capability for filtering and preserving high-quality solutions, thereby deepening exploration of the search space. In summary, TEMOF prevents premature convergence to local optima while facilitating the retention of diverse solution sets on the Pareto front. Ultimately, this contributes to the discovery of a more comprehensive set of Pareto optimal solutions.

\section{Conclusion and Future Works}
\label{Conclusion}
This paper introduces a Two-stage framework designed to enhance the performance of MOEAs. TEMOF achieves this by making simple adjustments to the structure of PMOEAs, duplicating the algorithm into two stages based on its original framework. While the two stages differ in the way offspring are generated, they remain consistent in other aspects. The first stage generates offspring solely from the current population, focusing on diversity search. The second stage involves collaborative search using an augmented external archive and the current population, emphasizing convergence search. The external archive is solely used to preserve the PF members at the first level. Subsequently, the framework is integrated with three PMOEAs. The winners are then compared with both MOEAs and PMOEAs, validating the effectiveness of the framework.

In future works, it is essential to design more advanced environmental selection methods to enhance the efficiency of filtering non-dominated solutions in the archive and improve the retention capability of high-quality non-dominated solutions. Additionally, efforts should be made to reduce the algorithm's dependence of the second stage on the first stage. This involves enhancing the global diversity search capability in the first stage, as well as improving the local diversity search capability and global convergence search capability in the second stage.

\ifCLASSOPTIONcaptionsoff
  \newpage
\fi

\scriptsize
\bibliographystyle{IEEEtran}
\bibliography{IEEEabrv,main}

\begin{thebibliography}{10}
\providecommand{\url}[1]{#1}
\csname url@samestyle\endcsname
\providecommand{\newblock}{\relax}
\providecommand{\bibinfo}[2]{#2}
\providecommand{\BIBentrySTDinterwordspacing}{\spaceskip=0pt\relax}
\providecommand{\BIBentryALTinterwordstretchfactor}{4}
\providecommand{\BIBentryALTinterwordspacing}{\spaceskip=\fontdimen2\font plus
\BIBentryALTinterwordstretchfactor\fontdimen3\font minus \fontdimen4\font\relax}
\providecommand{\BIBforeignlanguage}[2]{{%
\expandafter\ifx\csname l@#1\endcsname\relax
\typeout{** WARNING: IEEEtran.bst: No hyphenation pattern has been}%
\typeout{** loaded for the language `#1'. Using the pattern for}%
\typeout{** the default language instead.}%
\else
\language=\csname l@#1\endcsname
\fi
#2}}
\providecommand{\BIBdecl}{\relax}
\BIBdecl

\bibitem{deb2016multi}
K.~Deb, K.~Sindhya, and J.~Hakanen, ``Multi-objective optimization,'' in \emph{Decision sciences}.\hskip 1em plus 0.5em minus 0.4em\relax CRC Press, 2016, pp. 161--200.

\bibitem{pereira2022review}
J.~L.~J. Pereira, G.~A. Oliver, M.~B. Francisco, S.~S. Cunha~Jr, and G.~F. Gomes, ``A review of multi-objective optimization: methods and algorithms in mechanical engineering problems,'' \emph{Archives of Computational Methods in Engineering}, vol.~29, no.~4, pp. 2285--2308, 2022.

\bibitem{sharma2022comprehensive}
S.~Sharma and V.~Kumar, ``A comprehensive review on multi-objective optimization techniques: Past, present and future,'' \emph{Archives of Computational Methods in Engineering}, vol.~29, no.~7, pp. 5605--5633, 2022.

\bibitem{zhou2011multiobjective}
A.~Zhou, B.-Y. Qu, H.~Li, S.-Z. Zhao, P.~N. Suganthan, and Q.~Zhang, ``Multiobjective evolutionary algorithms: A survey of the state of the art,'' \emph{Swarm and evolutionary computation}, vol.~1, no.~1, pp. 32--49, 2011.

\bibitem{zhang2014efficient}
X.~Zhang, Y.~Tian, R.~Cheng, and Y.~Jin, ``An efficient approach to nondominated sorting for evolutionary multiobjective optimization,'' \emph{IEEE Transactions on Evolutionary Computation}, vol.~19, no.~2, pp. 201--213, 2014.

\bibitem{deb2002fast}
K.~Deb, A.~Pratap, S.~Agarwal, and T.~Meyarivan, ``A fast and elitist multiobjective genetic algorithm: Nsga-ii,'' \emph{IEEE transactions on evolutionary computation}, vol.~6, no.~2, pp. 182--197, 2002.

\bibitem{he2017radial}
C.~He, Y.~Tian, Y.~Jin, X.~Zhang, and L.~Pan, ``A radial space division based evolutionary algorithm for many-objective optimization,'' \emph{Applied Soft Computing}, vol.~61, pp. 603--621, 2017.

\bibitem{zhang2007moea}
Q.~Zhang and H.~Li, ``Moea/d: A multiobjective evolutionary algorithm based on decomposition,'' \emph{IEEE Transactions on evolutionary computation}, vol.~11, no.~6, pp. 712--731, 2007.

\bibitem{deb2013evolutionary}
K.~Deb and H.~Jain, ``An evolutionary many-objective optimization algorithm using reference-point-based nondominated sorting approach, part i: solving problems with box constraints,'' \emph{IEEE transactions on evolutionary computation}, vol.~18, no.~4, pp. 577--601, 2013.

\bibitem{while2006faster}
L.~While, P.~Hingston, L.~Barone, and S.~Huband, ``A faster algorithm for calculating hypervolume,'' \emph{IEEE transactions on evolutionary computation}, vol.~10, no.~1, pp. 29--38, 2006.

\bibitem{zhou2006combining}
A.~Zhou, Y.~Jin, Q.~Zhang, B.~Sendhoff, and E.~Tsang, ``Combining model-based and genetics-based offspring generation for multi-objective optimization using a convergence criterion,'' in \emph{2006 IEEE international conference on evolutionary computation}.\hskip 1em plus 0.5em minus 0.4em\relax IEEE, 2006, pp. 892--899.

\bibitem{trautmann2013r2}
H.~Trautmann, T.~Wagner, and D.~Brockhoff, ``R2-emoa: Focused multiobjective search using r2-indicator-based selection,'' in \emph{Learning and Intelligent Optimization: 7th International Conference, LION 7, Catania, Italy, January 7-11, 2013, Revised Selected Papers 7}.\hskip 1em plus 0.5em minus 0.4em\relax Springer, 2013, pp. 70--74.

\bibitem{bader2011hype}
J.~Bader and E.~Zitzler, ``Hype: An algorithm for fast hypervolume-based many-objective optimization,'' \emph{Evolutionary computation}, vol.~19, no.~1, pp. 45--76, 2011.

\bibitem{hernandez2015improved}
R.~Hern{\'a}ndez~G{\'o}mez and C.~A. Coello~Coello, ``Improved metaheuristic based on the r2 indicator for many-objective optimization,'' in \emph{Proceedings of the 2015 annual conference on genetic and evolutionary computation}, 2015, pp. 679--686.

\bibitem{tian2018guiding}
Y.~Tian, X.~Zhang, R.~Cheng, C.~He, and Y.~Jin, ``Guiding evolutionary multiobjective optimization with generic front modeling,'' \emph{IEEE transactions on cybernetics}, vol.~50, no.~3, pp. 1106--1119, 2018.

\bibitem{li2015pareto}
M.~Li, S.~Yang, and X.~Liu, ``Pareto or non-pareto: Bi-criterion evolution in multiobjective optimization,'' \emph{IEEE Transactions on Evolutionary Computation}, vol.~20, no.~5, pp. 645--665, 2015.

\bibitem{zhou2023evolutionary}
S.~Zhou, X.~Mo, Z.~Wang, Q.~Li, T.~Chen, Y.~Zheng, and W.~Sheng, ``An evolutionary algorithm with clustering-based selection strategies for multi-objective optimization,'' \emph{Information Sciences}, vol. 624, pp. 217--234, 2023.

\bibitem{li2014evolutionary}
K.~Li, K.~Deb, Q.~Zhang, and S.~Kwong, ``An evolutionary many-objective optimization algorithm based on dominance and decomposition,'' \emph{IEEE transactions on evolutionary computation}, vol.~19, no.~5, pp. 694--716, 2014.

\bibitem{das1998normal}
I.~Das and J.~E. Dennis, ``Normal-boundary intersection: A new method for generating the pareto surface in nonlinear multicriteria optimization problems,'' \emph{SIAM journal on optimization}, vol.~8, no.~3, pp. 631--657, 1998.

\bibitem{yang2013grid}
S.~Yang, M.~Li, X.~Liu, and J.~Zheng, ``A grid-based evolutionary algorithm for many-objective optimization,'' \emph{IEEE Transactions on Evolutionary Computation}, vol.~17, no.~5, pp. 721--736, 2013.

\bibitem{mohan2005evaluating}
M.~Mohan, K.~Deb, and S.~Mishra, ``Evaluating the edomination based multi-objective evolutionary algorithm for a quick computation of pareto-optimal solutions,'' \emph{Evolutionary Computation}, vol.~13, no.~4, pp. 501--525, 2005.

\bibitem{zhang2014knee}
X.~Zhang, Y.~Tian, and Y.~Jin, ``A knee point-driven evolutionary algorithm for many-objective optimization,'' \emph{IEEE Transactions on Evolutionary Computation}, vol.~19, no.~6, pp. 761--776, 2014.

\bibitem{wang2012preference}
R.~Wang, R.~C. Purshouse, and P.~J. Fleming, ``Preference-inspired coevolutionary algorithms for many-objective optimization,'' \emph{IEEE Transactions on Evolutionary Computation}, vol.~17, no.~4, pp. 474--494, 2012.

\bibitem{yen2013performance}
G.~G. Yen and Z.~He, ``Performance metric ensemble for multiobjective evolutionary algorithms,'' \emph{IEEE Transactions on Evolutionary Computation}, vol.~18, no.~1, pp. 131--144, 2013.

\bibitem{tian2020coevolutionary}
Y.~Tian, T.~Zhang, J.~Xiao, X.~Zhang, and Y.~Jin, ``A coevolutionary framework for constrained multiobjective optimization problems,'' \emph{IEEE Transactions on Evolutionary Computation}, vol.~25, no.~1, pp. 102--116, 2020.

\bibitem{ming2021dual}
M.~Ming, A.~Trivedi, R.~Wang, D.~Srinivasan, and T.~Zhang, ``A dual-population-based evolutionary algorithm for constrained multiobjective optimization,'' \emph{IEEE Transactions on Evolutionary Computation}, vol.~25, no.~4, pp. 739--753, 2021.

\bibitem{liu2019handling}
Y.~Liu, H.~Ishibuchi, G.~G. Yen, Y.~Nojima, and N.~Masuyama, ``Handling imbalance between convergence and diversity in the decision space in evolutionary multimodal multiobjective optimization,'' \emph{IEEE Transactions on Evolutionary Computation}, vol.~24, no.~3, pp. 551--565, 2019.

\bibitem{li2021weighted}
W.~Li, T.~Zhang, R.~Wang, and H.~Ishibuchi, ``Weighted indicator-based evolutionary algorithm for multimodal multiobjective optimization,'' \emph{IEEE Transactions on Evolutionary Computation}, vol.~25, no.~6, pp. 1064--1078, 2021.

\bibitem{zhan2013multiple}
Z.-H. Zhan, J.~Li, J.~Cao, J.~Zhang, H.~S.-H. Chung, and Y.-H. Shi, ``Multiple populations for multiple objectives: A coevolutionary technique for solving multiobjective optimization problems,'' \emph{IEEE transactions on cybernetics}, vol.~43, no.~2, pp. 445--463, 2013.

\bibitem{wang2015cooperative}
J.~Wang, W.~Zhang, and J.~Zhang, ``Cooperative differential evolution with multiple populations for multiobjective optimization,'' \emph{IEEE Transactions on Cybernetics}, vol.~46, no.~12, pp. 2848--2861, 2015.

\bibitem{cui2016adaptive}
L.~Cui, G.~Li, Q.~Lin, J.~Chen, and N.~Lu, ``Adaptive differential evolution algorithm with novel mutation strategies in multiple sub-populations,'' \emph{Computers \& Operations Research}, vol.~67, pp. 155--173, 2016.

\bibitem{tian2016multi}
Y.~Tian, X.~Zhang, R.~Cheng, and Y.~Jin, ``A multi-objective evolutionary algorithm based on an enhanced inverted generational distance metric,'' in \emph{2016 IEEE congress on evolutionary computation (CEC)}.\hskip 1em plus 0.5em minus 0.4em\relax IEEE, 2016, pp. 5222--5229.

\bibitem{van1998multiobjective}
D.~A. Van~Veldhuizen and G.~B. Lamont, ``Multiobjective evolutionary algorithm research: A history and analysis,'' Technical Report TR-98-03, Department of Electrical and Computer Engineering~…, Tech. Rep., 1998.

\bibitem{price2023trial}
K.~V. Price, A.~Kumar, and P.~Suganthan, ``Trial-based dominance for comparing both the speed and accuracy of stochastic optimizers with standard non-parametric tests,'' \emph{Swarm and Evolutionary Computation}, vol.~78, p. 101287, 2023.

\bibitem{li2019comparison}
H.~Li, K.~Deb, Q.~Zhang, P.~N. Suganthan, and L.~Chen, ``Comparison between moea/d and nsga-iii on a set of novel many and multi-objective benchmark problems with challenging difficulties,'' \emph{Swarm and Evolutionary Computation}, vol.~46, pp. 104--117, 2019.

\bibitem{tian2017indicator}
Y.~Tian, R.~Cheng, X.~Zhang, F.~Cheng, and Y.~Jin, ``An indicator-based multiobjective evolutionary algorithm with reference point adaptation for better versatility,'' \emph{IEEE Transactions on Evolutionary Computation}, vol.~22, no.~4, pp. 609--622, 2017.

\bibitem{moreira2019guiding}
G.~Moreira and L.~Paquete, ``Guiding under uniformity measure in the decision space,'' in \emph{2019 IEEE Latin American Conference on Computational Intelligence (LA-CCI)}.\hskip 1em plus 0.5em minus 0.4em\relax IEEE, 2019, pp. 1--6.

\bibitem{jiang2017strength}
S.~Jiang and S.~Yang, ``A strength pareto evolutionary algorithm based on reference direction for multiobjective and many-objective optimization,'' \emph{IEEE Transactions on Evolutionary Computation}, vol.~21, no.~3, pp. 329--346, 2017.

\bibitem{tian2017platemo}
Y.~Tian, R.~Cheng, X.~Zhang, and Y.~Jin, ``Platemo: A matlab platform for evolutionary multi-objective optimization [educational forum],'' \emph{IEEE Computational Intelligence Magazine}, vol.~12, no.~4, pp. 73--87, 2017.

\end{thebibliography}
\thispagestyle{empty}

\end{document}